\documentclass[sort&compress, numafflabel]{elsarticle}

\usepackage[]{natbib}
\usepackage[breaklinks,hidelinks]{hyperref}
\usepackage{times}
\usepackage{latexsym}

\usepackage{geometry}
\usepackage{subfiles}
\usepackage{multirow}
\usepackage{array}
\usepackage{hyphenat}
\usepackage{subcaption}
\usepackage{longtable}
\usepackage{graphicx}
\usepackage{adjustbox}
\usepackage{enumitem}
\usepackage{url}

\usepackage[color=yellow, textsize=small]{todonotes}
\usepackage{bm}
\usepackage{booktabs}
\usepackage{float}
\usepackage[english]{babel}
\usepackage{blindtext}
\usepackage{textcomp}
\usepackage{soul,color}
\usepackage{pifont}

\makeatletter
\def\ps@pprintTitle{%
 \let\@oddhead\@empty
 \let\@evenhead\@empty
 \def\@oddfoot{}%
 \let\@evenfoot\@oddfoot}
\makeatother
\usepackage{epstopdf}
\AppendGraphicsExtensions{.tif}

\usepackage{titlesec}
\titleformat{\section}
      {\normalfont\bfseries}
      {\thesection}
      {0ex}
      {\MakeUppercase}
\titleformat{\subsection}
      {\normalfont\bfseries}
      {\thesection}
      {0ex}
      {}
\titleformat{\subsubsection}
      {\normalfont}
      {\thesection}
      {0ex}
      {}

\newcommand\blfootnote[1]{%
  \begingroup
  \renewcommand\thefootnote{}\footnote{#1}%
  \addtocounter{footnote}{-1}%
  \endgroup
}

\usepackage{setspace}
\newif\ifsubfile
\subfiletrue
\newif\iftif
\tiffalse

\usepackage[utf8]{inputenc}
\usepackage[T1]{fontenc}
\usepackage{lineno}
\usepackage[numbers]{natbib}
\usepackage{subfiles}
\usepackage{graphicx}
\usepackage{float}
\usepackage{hyperref}
\usepackage{mathtools}
\usepackage{changepage}
\usepackage{caption}
\usepackage{multirow}
\usepackage{titlesec}
\usepackage{amsmath}

\usepackage{makecell}

\usepackage[math]{cellspace}
\usepackage{tikz}

\usepackage{url}

\usepackage{colortbl}
\linenumbers
\nolinenumbers

\usepackage{listings}
\usepackage{xcolor}

\title{LeafAI: query generator for clinical cohort discovery \\ rivaling a human programmer}

\author[bime,rit]{Nicholas J Dobbins\textsuperscript{+}\blfootnote{\textsuperscript{+}Corresponding author: Nicholas Dobbins, PhD, MLIS, Department of Biomedical Informatics and Medical Education, University of Washington, 1851 NE Grant Ln, Seattle, WA 98195, USA; ndobb@uw.edu.}}

\ead{}
\author[ischool]{Bin Han}\ead{}
\author[bime]{Weipeng Zhou}\ead{}
\author[med]{Kristine F Lan}\ead{}
\author[med]{H. Nina Kim}\ead{}
\author[med]{Robert Harrington}\ead{}
\author[ist]{Özlem Uzuner}\ead{}
\author[bime]{Meliha Yetisgen}\ead{}

\address[bime]{Department of Biomedical Informatics \& Medical Education, University of Washington, Seattle, WA, USA}
\address[rit]{Department of Research IT, UW Medicine, University of Washington, Seattle, WA, USA}
\address[med]{Department of Medicine, University of Washington, Seattle, WA}
\address[ischool]{Information School, University of Washington, Seattle, WA}
\address[ist]{Department of Information Sciences and Technology, George Mason University, Fairfax, VA \blfootnote{ \\ Word count: 4655} }

\begin{document}
\subfilefalse

\newpageafter{author}

\begin{abstract}

\noindent\textbf{Objective:} Identifying study-eligible patients within clinical databases is a critical step in clinical research. However, accurate query design typically requires extensive technical and biomedical expertise. We sought to create a system capable of generating data model-agnostic queries while also providing novel logical reasoning capabilities for complex clinical trial eligibility criteria. \\

\noindent\textbf{Materials and Methods:} The task of query creation from eligibility criteria requires solving several text-processing problems, including named entity recognition and relation extraction, sequence-to-sequence transformation, normalization, and reasoning. We incorporated hybrid deep learning and rule-based modules for these, as well as a knowledge base of the Unified Medical Language System (UMLS) and linked ontologies. To enable data-model agnostic query creation, we introduce a novel method for tagging database schema elements using UMLS concepts. To evaluate our system, called LeafAI, we compared the capability of LeafAI to a human database programmer to identify patients who had been enrolled in 8 clinical trials conducted at our institution. We measured performance by the number of actual enrolled patients matched by generated queries. \\

\noindent\textbf{Results:} LeafAI matched a mean 43\% of enrolled patients with 27,225 eligible across 8 clinical trials, compared to 27\% matched and 14,587 eligible in queries by a human database programmer. The human programmer spent 26 total hours crafting queries compared to several minutes by LeafAI.  \\

\noindent\textbf{Conclusions:} Our work contributes a state-of-the-art data model-agnostic query generation system capable of conditional reasoning using a knowledge base. We demonstrate that LeafAI can rival an experienced human programmer in finding patients eligible for clinical trials.

\end{abstract}


\maketitle

\pagebreak

\section*{Introduction}
\label{sec:background}

Identifying groups of patients meeting a given set of eligibility criteria is a critical step for recruitment into randomized controlled trials (RCTs). Often, clinical studies  fall short of recruitment goals, leading to time and cost overruns or  challenges in ensuring adequate statistical power \cite{gul2010clinical, adams2015barriers}. Failure to recruit research subjects may result from a variety of factors, but often stems from difficulties in translating complex eligibility criteria into effective queries that can sift through data  in the electronic health record (EHR) \cite{wang2017classifying}. Despite these difficulties, RCT investigators  increasingly rely on EHR data queries to  identify research candidates instead of labor-intensive manual chart or case report form review \cite{cowie2017electronic}. At the same time, the amount and variety of data contained in EHRs is increasing dramatically, creating both challenges and opportunities for patient recruitment \cite{lee2017medical}. While more granular and potentially useful data are captured and stored in EHRs now than in the past, the process of accessing and leveraging that data requires technical expertise and extensive knowledge of biomedical terminologies and data models. 

Cohort discovery tools such as Leaf \cite{dobbins2019leaf} and i2b2 \cite{murphy2010serving} may be used in many situations, as they offer relatively simple drag-and-drop interfaces capable of querying EHR data to find patients meeting given criteria \cite{johnson2014use}.  However, these tools have limitations, since their use often requires significant training and the tools have difficulty representing particularly complex nested or temporal eligibility criteria \cite{deshmukh2009evaluating}. Moreover, existing cohort discovery tools lack functionality to dynamically reason upon non-specific criteria that frequently appear in real-world eligibility criteria. For example, a criterion may require patients "indicated for bariatric surgery", but translating such non-specific criteria into a query (e.g., patients with a diagnosis of morbid obesity or body mass index greater than 40) must be performed manually by a researcher, even in cases where constructing an exhaustive list of such criteria may be time-intensive, subjective, and error-prone. 

In recent years, alternatives to web-based cohort discovery tools have been explored. In particular, various methods using Natural Language Processing (NLP) have been put forth by the research community \cite{yuan2019criteria2query, soni2020patient, fang2022combining, zhang2020deepenroll, chen2019clinical, patrao2015recruit, dhayne2021emr2vec, liu2021evaluating, xiong2019cohort}. NLP-based cohort discovery methods could be especially valuable since they can key on eligibility criteria described in natural language, a medium that clinicians, researchers and investigators already use.

\section*{Background and Significance}
\label{sec:background}

Various methods for cohort discovery using NLP have been previously explored. Yuan \textit{et al} developed Criteria2Query \cite{yuan2019criteria2query}, a hybrid information extraction (IE) pipeline and application which uses both rules and machine learning to generate database queries on an Observation Medical Outcomes Partnership (OMOP) \cite{hripcsak2015observational} database, later expanded by Fang \textit{et al} \cite{fang2022combining}. Other research has used encoder-decoder neural architectures for transforming clinical natural language questions into SQL queries \cite{bae2021question, park2021knowledge, wang2020text, pan2021bert, dhayne2021emr2vec}. These studies include exploration of cross-domain transformations, where systems must generalize to unseen database schema \cite{park2021knowledge}, handling of typos and abbreviations \cite{bae2021question}, and the generation of intermediate representations between a natural language utterance and final SQL database query.\cite{pan2021bert} 

Beyond database query generation, other cohort discovery methods explored include document ranking and classification \cite{chen2019clinical,soni2020patient} where clinical notes are summarized, ranked and classified as relevant to a given eligibility criterion, and embedding projections for entailment prediction \cite{dhayne2021emr2vec, zhang2020deepenroll} where predicting that a patient can be inferred from a given eligibility criteria equates to eligibility. Other studies have explored the use of ontologies and OWL-based reasoning in determining eligibility \cite{patel2007matching, huang2013semanticct, baader2018patient, johnson2016mimic, patrao2015recruit}.
    
\subsection*{Gaps and opportunities}

To date, most programs capable of generating database queries do so for only a single database schema, such as OMOP or MIMIC-III \cite{johnson2016mimic}. This lack of flexibility limits their capability to accommodate real-world project needs \cite{belenkaya2021extending, peng2021towards, zoch2021adaption, warner2019hemonc, zhou2013evaluation, shin2019genomic, kwon2019development}, such as adding new database tables to OMOP for cancer staging \cite{belenkaya2021extending}. Moreover, most methods, particularly those using direct text-to-SQL deep learning approaches, tend to generate relatively simple SQL statements with few JOINs or nested sub-queries and typically no support for UNION operators and so on. This relative simplicity contrasts with the complexity of real-world EHR databases, which may contain dozens or even hundreds of tables using various vocabularies and mappings. Furthermore, direct text-to-SQL methods are bound to SQL syntax, and thus incapable of querying other systems such as Fast Healthcare Interoperability Resources (FHIR) \cite{bender2013hl7}. Additionally, few of the methods described provide support for complex logic such as nested Boolean statements or temporal sequences, and none support reasoning on non-specific criteria (e.g., "diseases that affect respiratory function"), phenomena common to study eligibility criteria \cite{wang2017classifying, ross2010analysis}. Perhaps most importantly, to the best of our knowledge, only one previous work has been tested in terms of matching patients actually enrolled in clinical trials \cite{zhang2020deepenroll}, and none have been directly compared to the capabilities of a human database programmer.

\subsection*{Key Contributions}

We introduce the LeafAI query engine, an application capable of generating database queries for cohort discovery from free-text eligibility criteria. This work contributes the following:

\begin{enumerate}
    \item{A novel database annotation schema and mapping method to enable \textbf{data model-agnostic} query generation from natural language.}
    \item{Methods for transforming and leveraging \textbf{intermediate logical representations} of eligibility criteria.}
    \item{A \textbf{corpus of human-annotated logical forms of eligibility criteria} available to the research community\footnote{Will be made available upon article acceptance}.}
    \item{Methods for dynamically \textbf{reasoning upon non-specific criteria} using an integrated knowledge base (KB) of biomedical concepts.}
    \item{An evaluation of system performance by \textbf{direct comparison to that of a human database programmer} on actual clinical trial enrollments}.
\end{enumerate}

\section*{Materials and Methods}
\label{sec:methods}

\subsection*{System Architecture}

\begin{figure}[h]
  \includegraphics[scale=0.45]{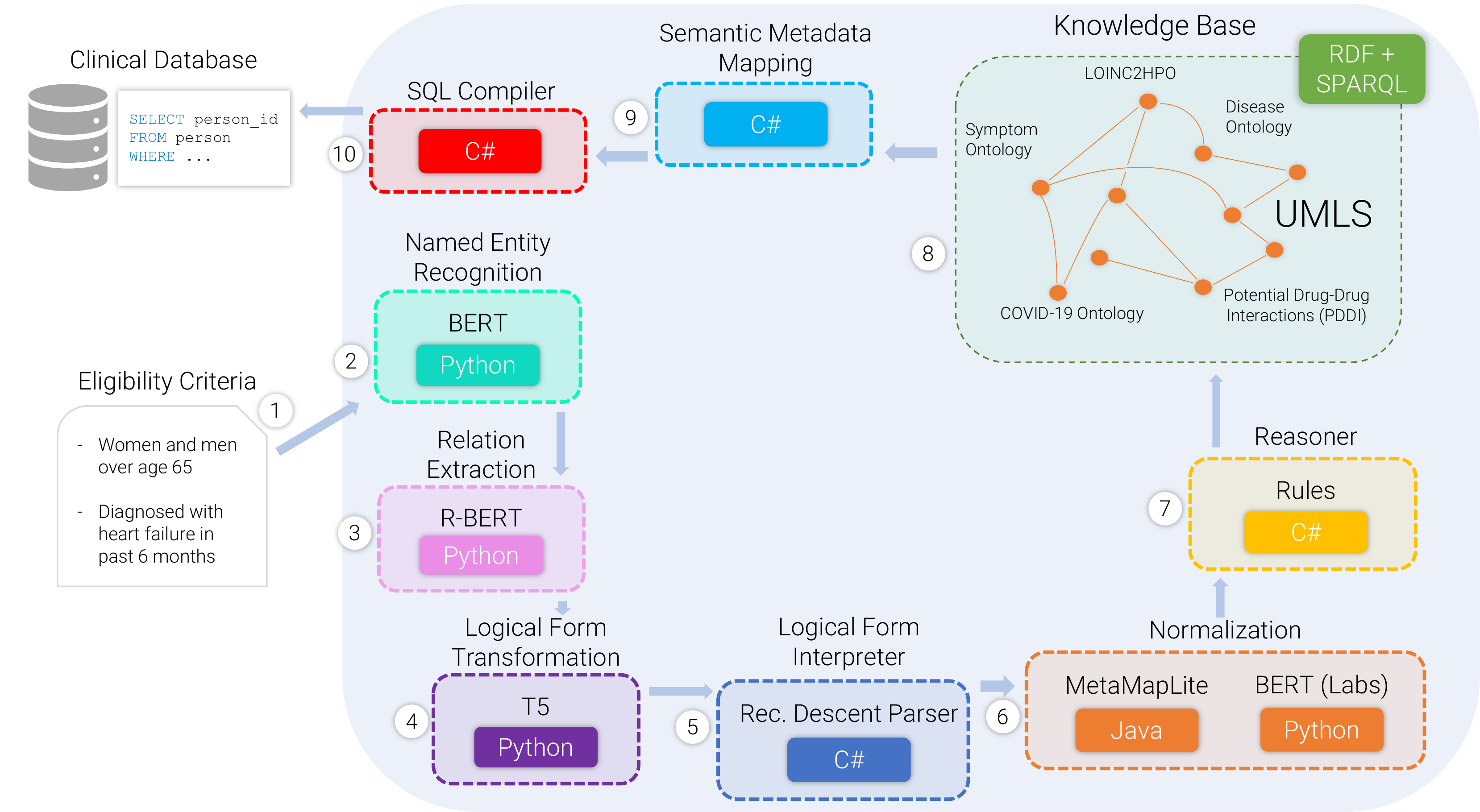}  
\caption{LeafAI query architecture. Inter-module communication is performed using the gRPC framework. Individual modules are deployed as Docker \cite{docker} containers and communicate solely with the central API, which orchestrates query generation and handles query generation requests.}
\label{fig_leafai_architecture}
\end{figure}

The LeafAI query engine was designed using a modular, micro service-based architecture with a central Application Program Interface (API) which orchestrates end-to-end query generation. Inter-module communication is performed using gRPC \cite{grpc}, a robust open-source remote procedure call framework which enables language-agnostic service integration. This allows individual modules to be implemented (and substituted) in programming languages and using libraries well-suited to a given task. A diagram of the LeafAI query engine architecture is shown in Figure \ref{fig_leafai_architecture}. 

At a high level, query generation is performed in the following steps:

\begin{enumerate}
    \item{A query request is received by the API in the form of inclusion and exclusion criteria as free-text strings.}
    \item{The input texts are tokenized and named entity recognition is performed to determine spans of text representing conditions, procedures, and so on.}
    \item{Relation extraction is performed to determine relations between the entities, such as \textit{Caused-By} or \textit{Numeric-Filter}.}
    \item{The input texts are transformed by replacing spans of "raw" text with logical form names. For example, "Diagnosed with diabetes" would become "Diagnosed with cond("diabetes")." The resulting input texts are in turn transformed into an output logical representation using a Sequence to Sequence (Seq2Seq) architecture, in the form of a string.}
    \item{A logical form interpreter module implemented as a recursive descent parser \cite{johnstone1998generalised} reads the logical form string input and instantiates it as an abstract syntax tree (AST) of nested in-memory logical form objects.}
    \item{"Named" logical form objects (i.e., specified with quoted text, such as "cond("diabetes")") are normalized into one or more corresponding UMLS concepts. UMLS child concepts are also added using our KB. For example, "cond("type 2 diabetes")” would also include concepts for type 2 diabetes with kidney complications (C2874072).}
    \item{Working recursively inside-to-outside the AST structure, each logical form object calls a \textit{Reason()} method which executes various rules depending on context.}
    \item{Each reasoning rule is performed as one or more pre-defined SPARQL queries to the KB, concept by concept.}
    \item{The final normalized, reasoned, logical form AST is thus a nested structure of UMLS concepts. Each AST criterion is mapped to zero or more corresponding entries in the semantic metadata mapping (SMM), which in turn lists meanings, roles, and relations of a database schema in the form of UMLS concepts.}
    \item{The final mapped AST object is transformed into a series of database queries, one per line of eligibility criteria text. The output SQL query can either be executed directly on a database or returned to the API caller.}
\end{enumerate}

\begin{figure}[h]
  \includegraphics[scale=0.44]{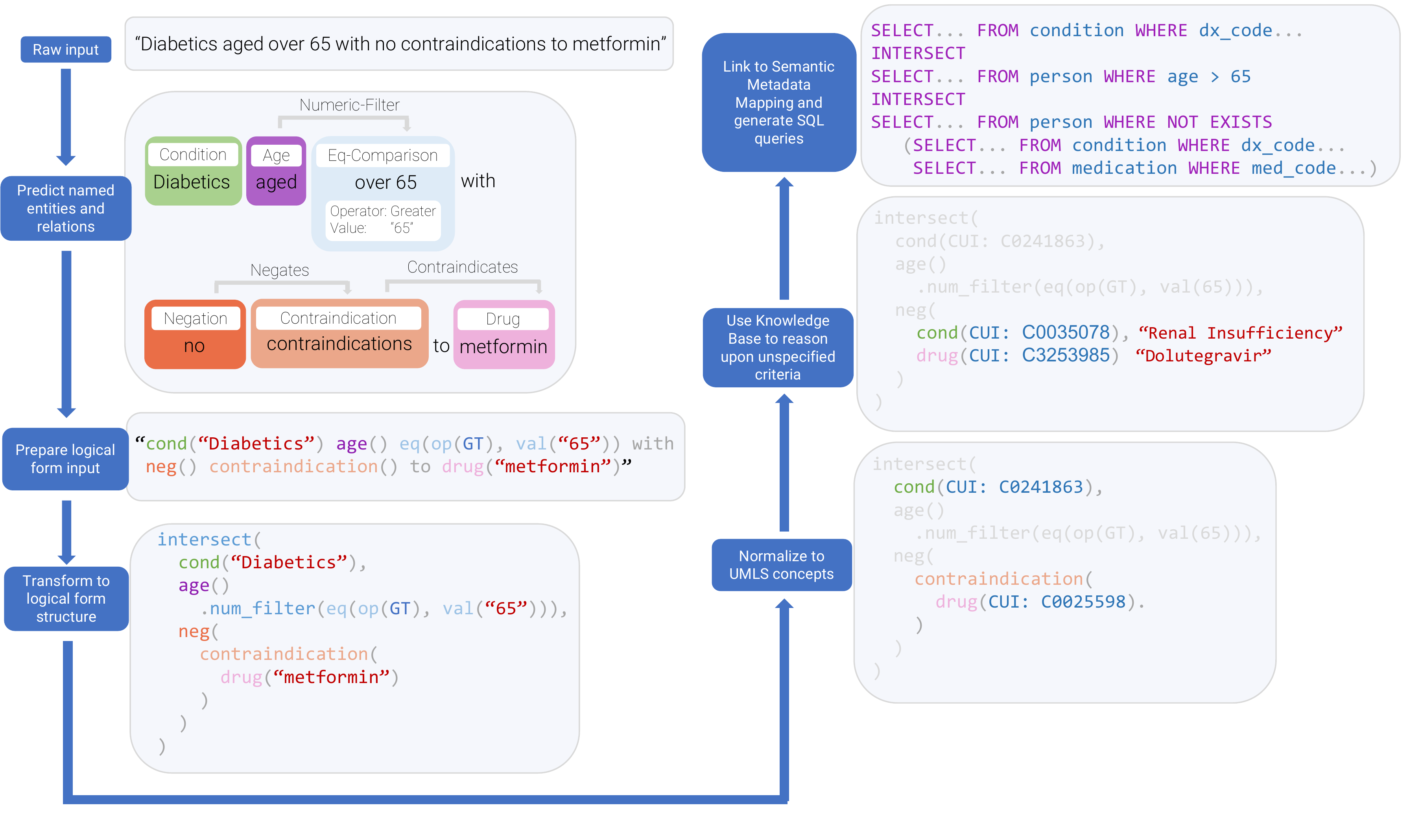}  
\caption{LeafAI query generation processes}
\label{fig_leafai_querygen}
\end{figure}

\noindent Figure \ref{fig_leafai_querygen} illustrates an example of this process. In the following subsections we examine these steps in detail.

\subsection*{Named entity recognition and relation extraction}

\noindent Named entity recognition (NER) refers to the segmentation and identification of tokens within an input sentence as "entities", such as conditions or procedures. We used the Leaf Clinical Trials (LCT) corpus \cite{dobbins2022leaf} to train two BERT-based \cite{devlin2018bert} NER extractors, one each for LCT general- and fine-grained-entities (see \cite{dobbins2022leaf} for more information on LCT entity types). Next, we perform relation extraction between named entity pairs similarly using a BERT-based model also trained on the LCT corpus.

\subsection*{Logical form transformation}

\noindent One of the core challenges of generating queries for eligibility criteria is the problem of logical representation. Generating queries directly based on named entities and relations alone, while practical, performs poorly in cases of nested or particularly complex logic. An alternative to this approach is to use a so-called intermediate representation (IR), which transforms the original natural language by removing "noise" unnecessary to a given task and which more logically represents underlying semantics (see Herzig \textit{et al} \cite{herzig2021unlocking} for an examination of IR-based SQL generation approaches). Similar to earlier work using Description Logics, Roberts and Demner-Fushman \cite{roberts2016annotating} proposed a representation of questions on EHR databases using a comparatively compact but flexible format using first order logic expressions, for example, representing "Is she wheezing this morning?" as

\begin{quote}
    \centering
    $\delta( \lambda x.has\_problem(x, C0043144, status) \wedge time\_within(x, \mathrm{"this\ morning"}))$
\end{quote}

\noindent This style of representation is powerfully generalizable, but also difficult to translate directly into SQL statements as multiple predicates (e.g., \textit{has\_problem} and \textit{time\_within}) may actually correspond to one or many SQL statements, depending on context, complicating direct transformation into queries.

We thus chose a similar intermediate representation (hereafter simply "logical forms") as proposed by Roberts and Demner-Fushman but more closely resembling a nested functional structure in programming languages such as Python or JavaScript and more amenable to SQL generation. A criterion such as "Diabetic women and men over age 65" would be represented by our logical forms as

\begin{quote}
$intersect( \\
    \mathrm{\ \ \ \ }cond("Diabetic"), \\
    \mathrm{\ \ \ \ }union(female(), male()),\\
    \mathrm{\ \ \ \ }age().num\_filter(eq(op(GT), val("65"))) \\
)$
\end{quote}

\noindent A description of our logical forms annotation schema, corpus, called the Leaf Logical Forms (LLF) corpus, annotation process, and performance metrics can be found in Appendix A.

After NER and relation extraction are performed, we leverage T5 \cite{raffel2020exploring}, a state-of-the-art Seq2Seq architecture we fine-tuned for predicting logical forms on the LLF corpus. As inputs to the Seq2Seq model we use the original eligibility criteria with named entity spans replaced by logical form representations, since we found this improved performance compared to training with raw inputs. Thus the above example input would be transformed to

\begin{quote}
    \centering
    $\textit{"cond(“Diabetic”) female() and male() over age() eq(op(GT), val(“65”))"}$
\end{quote}

\noindent The returned logical form string is then instantiated into an abstract syntax tree (AST) of nested in-memory logical form objects using a recursive descent parser \cite{johnstone1998generalised} within our API.

\subsection*{Concept normalization}

Normalization refers to the mapping of free-text string values (e.g., "diabetes mellitus") to coded representations (e.g., UMLS, ICD-10, SNOMED, or LOINC). We normalize "named" logical forms to UMLS concepts using MetaMapLite \cite{aronson2001effective, demner2017metamap}. We consider a logical form "named" if it contains a free-text value surrounded by quotes. For example, \textit{cond()} is unnamed and refers to any condition or disease, while \textit{cond("hypertension")} is named as it refers to a specific condition. 

Normalization using MetaMapLite can often result in high recall but low precision, as MetaMapLite has no NER component and tends to return UMLS concepts which match a given phrase syntactically but refer to abstract concepts not of interest (e.g., a search for "BMI" may return "body mass index" (C1305855), but also organic chemical "BMI 60" (C0910133)). To improve normalization precision, we employ two strategies. First, our NER component filters predicted UMLS concepts to only those of specific semantic types. For example, we limit condition concepts to only those related to semantic types of diseases or syndrome (dsyn) and so on. Next, using term-frequencies pre-computed across UMLS concept phrases, we compare term frequency-inverse document frequency (tf-idf) on MetaMapLite predictions, removing UMLS concepts whose summed matched spans have a tf-idf score lower than that of unmatched spans in a given named entity. For example, for the string "covid-19 infection", MetaMapLite predicts both "COVID-19" (C5203670) as well as several concepts related to general infections. Our tf-idf strategy removes general infection concepts because “infection” has a lower tf-idf score than the summed scores for “covid” + “-” + “19”. 

Laboratory values present a particular challenge, as LeafAI expects predicted lab concepts to have directly associated LOINC codes, while MetaMapLite typically normalizes lab test strings to UMLS concepts of semantic type "laboratory test or finding", but which do not have direct mappings to LOINC codes. For example, a search for "platelet count" returns the concept "Platelet Count Measurement" (C0032181), but not the needed concept of "Platelet \# Bld Auto" (C0362994) with LOINC code “777-3”. Thus similar to Lee and Uzuner with medications \cite{lee2020normalizing}, we trained a BERT model for sequence classification to normalize lab tests. We trained this model to identify UMLS concepts associated with LOINC codes most frequently used in eligibility criteria \cite{rafee2022elapro}, with each CUI as a possible class.

\subsection*{Reasoning using an integrated knowledge base}

For reasoning and derivation of ICD-10, LOINC, and other codes for UMLS concepts, we designed a KB accessible via SPARQL queries and stored as Resource Description Framework (RDF) \cite{manola2004rdf} triples. The core of our KB is the UMLS, derived using a variation of techniques created for ontologies in BioPortal \cite{noy2009bioportal}. To further augment the UMLS, we mapped and integrated the Disease Ontology \cite{schriml2012disease}, Symptom Ontology \cite{sayers2010database}, COVID-19 Ontology \cite{sargsyan2020covid}, Potential Drug-Drug Interactions \cite{ayvaz2015toward}, LOINC2HPO \cite{zhang2019semantic}, and the Disease-Symptom Knowledge Base \cite{wang2008automated}, as well as ICD-9, ICD-10, SNOMED, and other general equivalence mappings not present in the UMLS \cite{}. We then developed SPARQL queries parameterized by UMLS concepts for various scenarios which leveraged our KB, such as contraindications to treatments, symptoms of diseases, and so on. Using LOINC2HPO mappings further allows us to infer phenotypes by lab test results rather than using ICD-10 or SNOMED codes alone. 

Our KB, nested logical forms, and inside-to-outside normalization methods enable "multi-hop" reasoning on eligibility criteria over several steps. For example, given the non-specific criterion "Contraindications to drugs for conditions which affect respiratory function", our system successfully reasons that (among other results),

\begin{enumerate}
    \item \textbf{Asthma} causes changes to \textbf{respiratory function}
    \item \textbf{Methylprednisolone} can be used to treat \textbf{asthma}
    \item \textbf{Mycosis} (fungal infection) is a contraindication to \textbf{methylprednisolone}
\end{enumerate}

\noindent These features allow LeafAI to reason upon fairly complex non-specific criteria.

\subsection*{Query generation using semantic metadata mapping}

\begin{figure}[h]
  \includegraphics[scale=0.47]{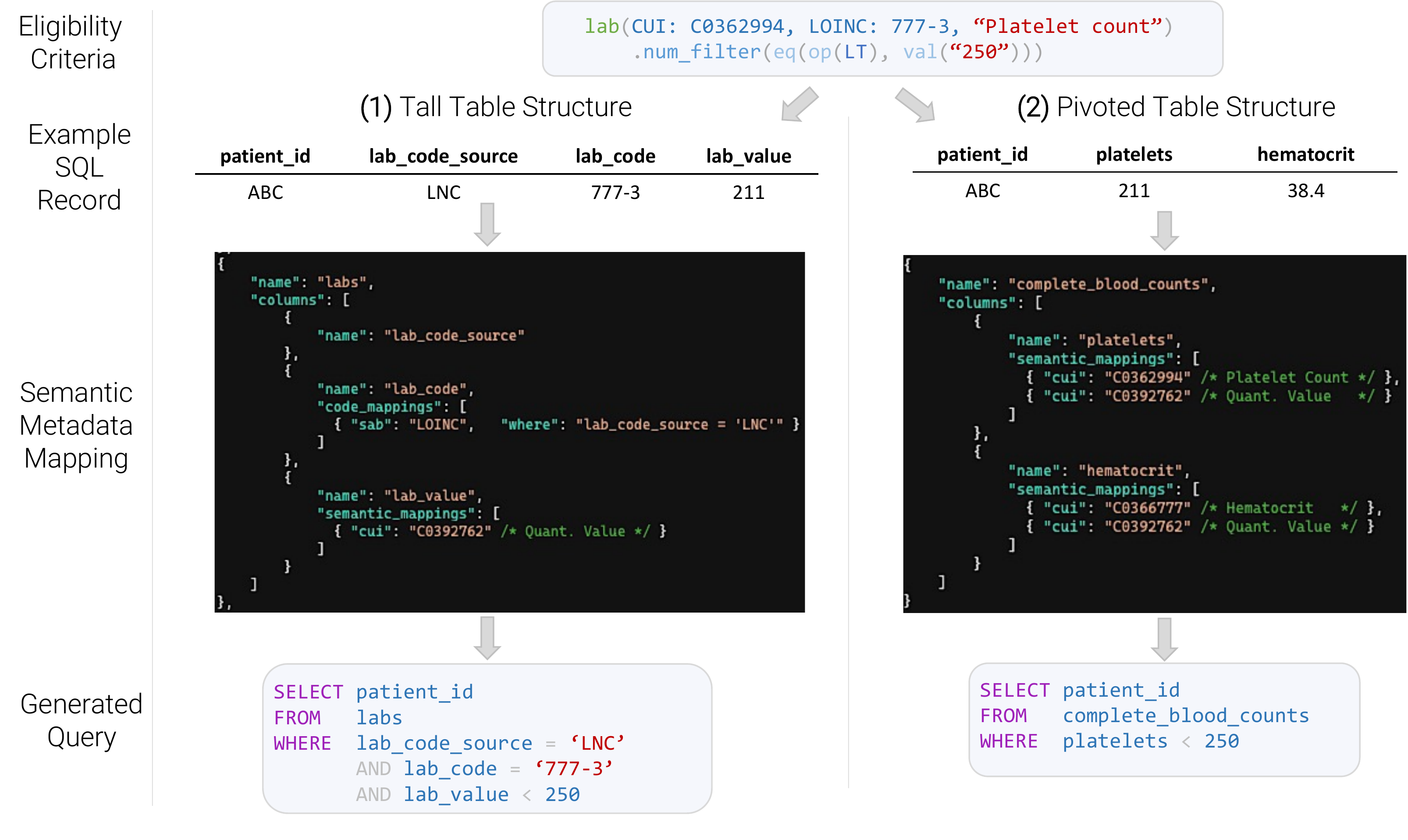}  
\caption{The LeafAI query engine's SQL query generation process using two hypothetical database schema to generate queries for platelet counts (shown in logical form after normalization). This example illustrates the flexibility of LeafAI's semantic metadata mapping system (represented here in JSON format) in adapting to virtually any data model. On the left, "Tall Table Structure", platelet counts must be filtered from within a general purpose "labs" table. The LeafAI KB recognizes that labs may be stored as LOINC codes, and the corresponding SMM indicates that records in this table can be filtered to LOINC values. On the right, "Pivoted Table Structure", platelet counts are stored as a specific column in a "complete\_blood\_counts" table, and thus can be directly queried without further filtering. Additional metadata, columns, tables, types and so on needed in SMMs are omitted for brevity.}
\label{fig_leafai_smm}
\end{figure}

To enable data model-agnostic query generation, we leveraged a subset of codes within the UMLS in what we define as a semantic metadata mapping, or SMM. An SMM includes a listing of available databases, tables, columns, and so on within a given database schema. Critically, these database artifacts are "tagged" using UMLS concepts. An example of this can be seen in Figure \ref{fig_leafai_smm}, which shows strategies by which a given criterion can be used to generate schema-specific queries by leveraging different SMMs. In cases where the LeafAI query engine finds more than one means of querying a concept (e.g., two SQL tables for diagnosis codes), the queries are combined in a UNION statement.

\subsection*{Evaluation}

It is reasonable to expect that an NLP-based system for finding patients based on eligibility criteria would find many  patients who actually enrolled in a real clinical trial — assuming that patients enrolled in those trials met the necessary criteria as determined by study investigators. While there are caveats to this approach (for example, certain diagnosis codes may be missing for some patients, etc.), we suggest that tools such as LeafAI be evaluated by their ability to handle real-world eligibility criteria and clinical data.

In this study we compared LeafAI’s results to that of a human database programmer with over 5 years of experience as a skilled analyst using SQL databases and other tools related to our EHR. This study was approved by the University of Washington’s Institutional Review Board. Our evaluation was performed as follows:

\begin{enumerate}
    \item We extracted metadata on 168 clinical trials from our EHR between January 2017 and December 2021 where at least 10 patients were indicated as enrolled and not withdrawn, and the total number of raw lines of free-text within the eligibility criteria (besides the phrases "Inclusion Criteria" and "Exclusion Criteria") was less than or equal to 30.
    \item By manual review, we excluded 22 trials with multiple sub-groups, as it would not be possible to know which eligibility criteria applied to which sub-group of enrolled patients.
    \item To narrow the scope of our evaluation, we chose to evaluate only trials studying the following 7 disease groups: Cardiology, COVID-19, Crohn's Disease, Multiple Sclerosis (MS), Diabetes Mellitus, Hepatitis C, and Cancer. Using the "condition" field for each trial within metadata from \url{https://clinicaltrials.gov}, we filtered and grouped the remaining 146 trials into only those studying our diseases of interest. These diseases were selected to provide a diverse representation of potential queries that not only span different systems or medical specialties (infection, malignancy, cardiac, neurologic, endocrinologic) but also included conditions (COVID-19, Crohn’s disease, diabetes) commonly studied in clinical trials.
    \item We randomly chose 1 trial from each group, with the exception of Cancer, where given the large number of trials and variety of cancer types, we chose 2 trials. 427 patients were enrolled across the chosen 8 clinical trials.
    \item Both LeafAI and the human programmer were provided the raw text of eligibility criteria from \url{https://clinicaltrials.gov}, (LeafAI using API calls). Each then created queries to find patients based on each eligibility criteria, which we executed on an OMOP database derived from our EHR containing our institution’s entire research-eligible patient population.
    \item To ensure results returned would be limited to only data available during the time of each trial, we replaced references to the SQL function for generating a current timestamp (\textit{GETDATE()}) with that of each trial's end date, and similarly replaced OMOP table references with SQL views filtering data to only that existing prior to the end of a trial.
    \item To ensure queries would be comparable to LeafAI, the human programmer was instructed to (1) ignore criteria which cannot be computed, such as “Willing and able to participate in the study” or criteria subject to “the opinion of the investigator”, as the intents and consents of patients and investigators are unavailable in our data, (2) make a best effort to reason upon non-specific criteria (e.g., symptoms for a condition), (3) not check whether patients found by a human query enrolled within a trial, and (4) skip criteria which cause an overall query to find no eligible patients. While our team did review examples of each of these cases, we did not define formal guidelines and the human programmer instead used their best judgment.
\end{enumerate}

\section*{Results}
\label{sec:results}

Results of the query generation experiment are shown in Table \ref{tbl_results}. Overall, LeafAI matched 212 of 427 (49\%) total enrolled patients across 8 clinical trials compared to 180 (42\%) found by queries of the human programmer. The mean per-trial percent of patients matched was 43.5\% for LeafAI and 27.2\% for the human programmer. LeafAI had a greater number of patients deemed eligible across all 8 trials, for a total of 27,225 eligible compared to 14,587 found by the human programmer. 

\begin{table}[h!]
    \small
    \centering
    \def\arraystretch{1.4}
\begin{tabular}{l l c c |r l |r l |c}
     & & & & \multicolumn{2}{c}{\textbf{LeafAI}} & \multicolumn{2}{|c}{\textbf{Human}}  \\
     \toprule
    \textbf{Condition} & \textbf{ID} & \textbf{\# Crit.} & \textbf{Enrolled} & \textbf{Matched} & \textbf{Eligible} & \textbf{Matched} & \textbf{Eligible} & \textbf{Time (hrs)} \\
    \midrule
    CL Lymphoma        & \footnotesize{NCT04852822} & 4 & 83 & 80 (96\%) & 3,252 & 77 (92\%) & 2,382 & 1 \\
    Hepatitis C        & \footnotesize{NCT02786537} & 8 & 42 & 33 (78\%) & 9,529 & 32 (76\%) & 9,372 & 4 \\
    Crohn's Disease    & \footnotesize{NCT03782376} & 9 & 16 & 0 (0\%) & 113 & 1 (6\%) & 9 & 2 \\
    Cardiac Arrest     & \footnotesize{NCT04217551} & 12 & 27 & 12 (44\%) & 4,792 & 0 (0\%) & 598 & 5 \\
    COVID-19           & \footnotesize{NCT04501952} & 13 & 41 & 0 (0\%) & 0 & 0 (0\%) & 98 & 2 \\
    Multiple Sclerosis & \footnotesize{NCT03621761} & 14 & 196 & 77 (39\%) & 4,891 & 69 (35\%) & 1,016 & 3 \\
    Type 1 Diabetes    & \footnotesize{NCT03335371} & 18 & 11  & 0 (0\%) & 1,006 & 1 (9\%) & 1,104 & 4 \\
    Ovarian Cancer     & \footnotesize{NCT03029611} & 25 & 11 & 10 (91\%) & 1,667 & 0 (0\%) & 8 & 5 \\
    \bottomrule
    \textbf{Mean} & & & & 43.5\% & & 27.2\% & \\
    \textbf{Total} & & 103 & 427 & 212 (49\%) & 27,225 & 180 (42\%) & 14,587 & 26  \\
\end{tabular}
    \caption{Statistics for each clinical trial evaluated by the LeafAI query engine and human programmer. The number of enrolled and matched patients were determined by cross-matching enrollments listed within our EHR. \textit{\# Crit.} indicates the number of lines of potential criteria, defined as any text besides blank spaces and the phrases “Inclusion criteria” and “Exclusion criteria”.}
    \label{tbl_results}
\end{table} 

Table \ref{tbl_results_leafai_detail} shows the number of criteria which were skipped by LeafAI. Of the 103 total criteria across all 8 studies, LeafAI executed queries for 56 (54.4\%) and skipped 5 (4.8\%) as it found no patients and 42 (40.7\%) because no computable concepts were found. 

\begin{table}[h!]
    \small
    \centering
    \def\arraystretch{1.4}
\begin{tabular}{l c |l l l| l l l}
     & & \multicolumn{3}{c}{\textbf{LeafAI}} & \multicolumn{3}{|c}{\textbf{Human}}  \\
    \toprule
    \textbf{Condition} & \textbf{\# Crit.} & \textbf{No Pats.} & \textbf{Not Computable} & \textbf{Executed} & \textbf{No Pats.} & \textbf{Not Computable} & \textbf{Executed} \\
    \toprule
    Cl Lymphoma        & 4  & 0 (0\%)    & 0 (0\%)   & 4  (100\%) & 0 (0\%)   & 1 (25\%)  & 3 (75\%) \\
    Hepatitis C        & 8  & 0 (0\%)    & 4 (50\%)  & 4  (50\%)  & 0 (0\%)   & 4 (50\%)  & 4 (50\%) \\
    Crohn's Disease    & 9  & 0 (0\%)    & 4 (44\%)  & 5  (55\%)  & 3 (33\%)  & 3 (33\%)  & 3 (33\%) \\
    Cardiac Arrest     & 12 & 0 (0\%)    & 8 (66\%)  & 4  (33\%)  & 2 (16\%)  & 5 (41\%)  & 5 (41\%) \\
    COVID-19           & 13 & 0 (0\%)    & 6 (46\%)  & 7  (53\%)  & 0 (0\%)   & 7 (53\%)  & 6 (46\%) \\
    Multiple Sclerosis & 14 & 1 (7\%)    & 3 (21\%)  & 10 (71\%)  & 1 (7\%)   & 5 (35\%)  & 8 (57\%) \\
    Type 1 Diabetes    & 18 & 2 (11\%)   & 8 (44\%)  & 8  (44\%)  & 4 (22\%)  & 6 (33\%)  & 8 (44\%) \\
    Ovarian Cancer     & 25 & 2 (8\%)    & 9 (36\%)  & 14 (56\%)  & 1 (4\%)   & 9 (36\%)  & 15 (60\%) \\
    \bottomrule
    \textbf{Total}     & 103 & 5 (5\%)   & 42 (41\%) & 56 (54\%)  & 11 (10\%) & 40 (39\%) & 52 (50\%) \\
\end{tabular}
    \caption{LeafAI and the human programmer’s handling of eligibility criteria for each trial. The column \textit{No Pats.} (Patients) indicates the count of criteria which would, if executed, cause no patients to be eligible. The column \textit{Not Computable} indicates the count of criteria which were non-computable, for various reasons. For both LeafAI and the human programmer these types of criteria were ignored.}
    \label{tbl_results_leafai_detail}
\end{table} 

Figure \ref{fig_leafai_results_analysis} shows differences in query strategies for 4 trials between LeafAI and the human programmer.

\section*{Discussion}
\label{sec:discussion}

Our results demonstrate that LeafAI is capable of rivaling the ability of a human programmer in identifying patients who are potentially eligible for  clinical trials. Indeed, in numerous cases we found LeafAI and the human programmer executing similar queries, such as for Hepatitis C (NCT04852822), Chronic Lymphocytic Leukemia (NCT04852822), MS (NCT03621761), and Diabetes Mellitus (NCT03029611), where both ultimately matched a similar number of patients. 243 unique patients were matched in total by either LeafAI or the human programmer, with 149 (61.2\%) identified by both.

One notable pattern we found is that LeafAI consistently finds a higher number of potentially eligible patients. We hypothesize that in many cases, LeafAI’s KB played a key role in this. For example, in the MS trial, LeafAI searched for 11 different SNOMED codes related to MS (including MS of the spinal cord, MS of the brain stem, acute relapsing MS, etc.), while the human programmer searched for only one, and ultimately LeafAI found nearly 5 times the number of potentially eligible patients (4,891 versus 1,016). It is possible that the human programmer had a lower rate of false positives (higher precision). However, this would come at the expense of manually reviewing tens of thousands of patient records to determine true eligibility and will be explored in a future analysis.  

\begin{figure}[H]
  \begin{center}
    \includegraphics[scale=0.51]{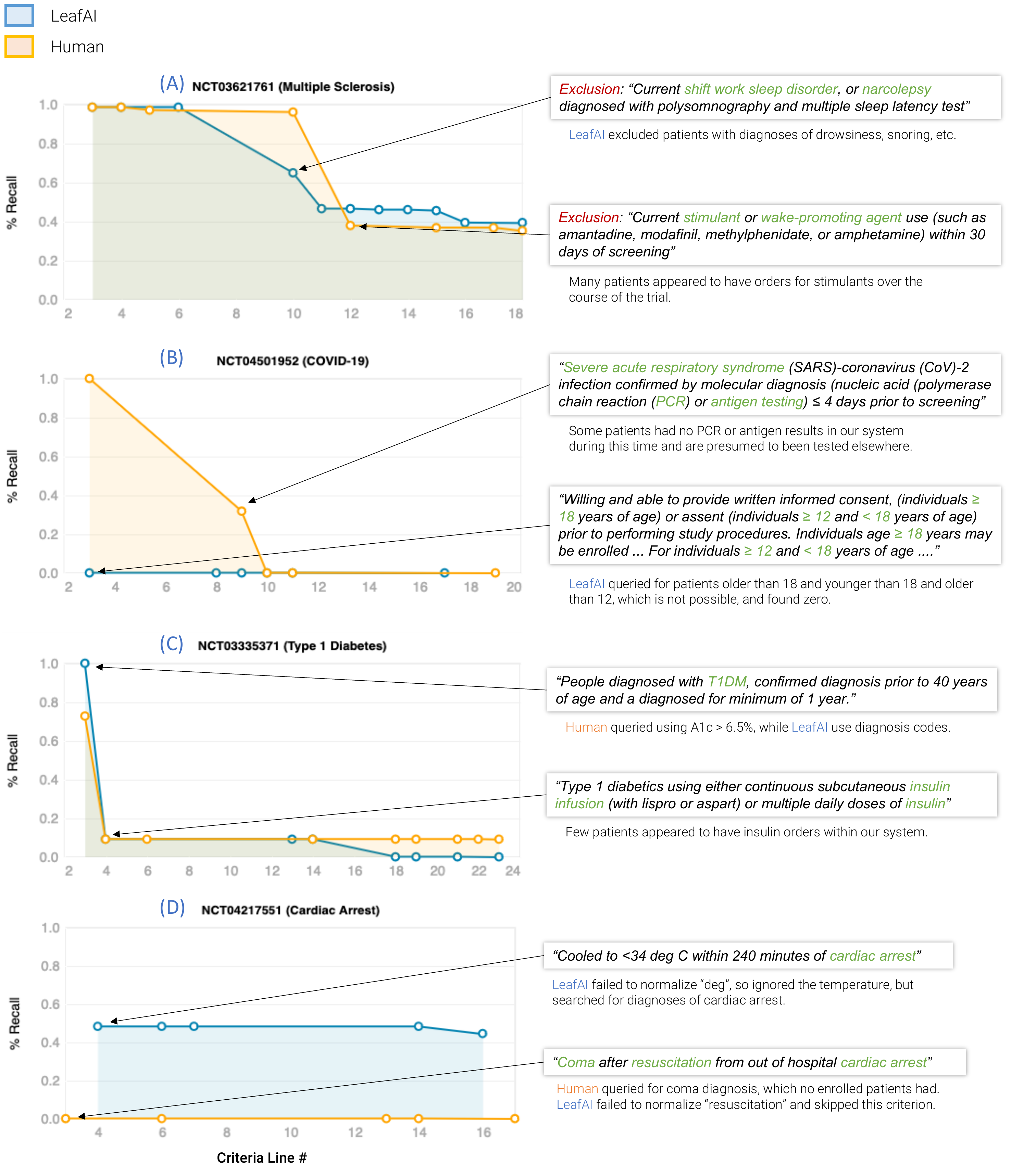}  
  \end{center}
  \caption{Longitudinal results listing patients found at each step in the query process for four trials illustrating data issues and differing query strategies between LeafAI and the human programmer. The blue line indicates recall for LeafAI and orange that of the human programmer. The X axis represents the line number within the free-text eligibility criteria. Dots indicate that a query was executed for a given line. On the right, boxes represent the text of a given eligibility criteria, with comments below discussing strategies and findings.}
  \label{fig_leafai_results_analysis}
\end{figure}

On the other hand, in the same trial, as can be seen in Figure \ref{fig_leafai_results_analysis} (A), given the exclusion criteria: "Current shift work sleep disorder, or narcolepsy diagnosed with polysomnography and multiple sleep latency", LeafAI’s KB unnecessarily excluded otherwise eligible patients by removing those with diagnosis codes for drowsiness, snoring, etc., since within the UMLS those are child concepts of sleep disorder (C0851578). The exclusion of these patients likely resulted in an approximately 40\% drop in recall at that stage compared to the human programmer, though ultimately both achieved similar recall (LeafAI: 39\% versus Human: 35\%).

Another challenging pattern we identified is the normalization of text to coded values. In the Ovarian Cancer trial (NCT03029611), both LeafAI and the human programmer matched eligible patients until line 10, which specified “Serum creatinine =< 2 or creatinine clearance > 60 ml/min...”. The human programmer was unable to find a LOINC code for creatinine clearance and instead queried only for serum creatinine, finding 3 relatively rare tests which none of the enrolled patients had performed. In contrast, LeafAI normalized the serum creatinine test to LOINC code 2160-0, which 10 patients had performed. In the case of the Cardiac Arrest trial (NCT04217551), as in Figure 4 (D), in the first criterion, “Coma after resuscitation from out of hospital cardiac arrest”, LeafAI attempted to create a temporal sequence query for coma diagnoses, but failed to normalize “resuscitation” and skipped the criterion as non-computable. The human programmer searched for patients with a coma diagnosis, which no enrolled patients had. In the following criterion, “Cooled to <34 deg C within 240 minutes of cardiac arrest”, LeafAI searched for patients with a diagnosis of cardiac arrest, which 12 patients had. LeafAI ultimately matched 12 of 27 (44\%) versus zero for the human programmer.  

Beyond performance measured by recall, it is notable that the human programmer spent approximately 26 hours crafting queries for the 8 trials while LeafAI took only several minutes running on a single laptop. The time saved by using automated means such as LeafAI for cohort discovery may save health organizations significant time and resources.

\subsection*{Limitations}

This project has some limitations. First, while the 8 clinical trials we evaluated were randomly selected, we specifically restricted the categories of diseases from which trials were chosen and limited to trials with 30 or less lines of eligibility criteria, and thus our results may not generalize to other kinds of trials. Next, we evaluated our queries using an OMOP-based extract which did not contain the full breadth of data within our EHR. Had our experiments been conducted using our enterprise data warehouse (populated by our EHR), it is possible the human programmer would have achieved better results than LeafAI due to knowledge and experience in utilizing granular source data. For example, in the Cardiac Arrest trial, the human programmer noted that data for use of cooling blankets is available in our EHR, but not in OMOP. It is not clear how LeafAI would perform were such data available. Our tests were also limited to only one institution, and it is possible that other institutions implementing different clinical trials and accessing different databases may find different results. We also did not directly compare LeafAI to other NLP systems. While we considered comparing LeafAI to Criteria2Query \cite{yuan2019criteria2query} as part of our baseline, our analysis reviewed results longitudinally (i.e., line by line of criteria), a function which Criteria2Query does not perform. Additionally, of the 103 criteria included in the 8 trials studied, LeafAI executed queries for only 56 (54\%) and the human for 52 (50\%) of them. While many criteria were unknowable (e.g., “In the opinion of investigators”) or not present in our data (e.g., “Consent to the study”), others were not computable due to failures of normalization or incorrectly predicted logical form structure. While the number of skipped criteria demonstrates that improvements to LeafAI are needed, the number of criteria found non-computable by the human programmer was similar, suggesting that potential improvements along these lines may be limited. Instead, study investigators might have to be more aware of computing shortfalls when designing criteria that will be executed by both programs and programmers; we leave a deeper analysis of this to future work. The number of truly eligible patients within our institution for each trial is unknown, which impedes our ability to measure system performance. We used each trial’s known enrolled patients as our gold standard, but assume they represent only a subset of those eligible. We recognize that additional analyses regarding the false positive and true negative rates are needed. These analyses were not undertaken in this study given our limited resources and the need for manual review of many thousands of patient records to complete them.

Last, we did not establish detailed guidelines for the human programmer when drafting queries, and possible inconsistencies in programmer interpretations of criteria may introduce subjective biases and challenges in replicating our results. On the other hand, we believe it is fair to assume an experienced clinical database programmer also brings background knowledge and intuition as to what kinds of data are captured, clinical workflows, relations between laboratory results and diagnosis codes, etc., which would be difficult or even counterproductive to attempt to encapsulate within specific guidelines. Additionally, as our team met and drafted basic (though not extensive) instructions for the human programmer to follow, the human programmer did have agreed-upon general parameters from which to interpret criteria. For these reasons we believe that allowing the human programmer leeway in interpreting criteria was a pragmatic and reasonable approach.

\subsection*{Future work}

\begin{figure}[H]
  \includegraphics[scale=0.26]{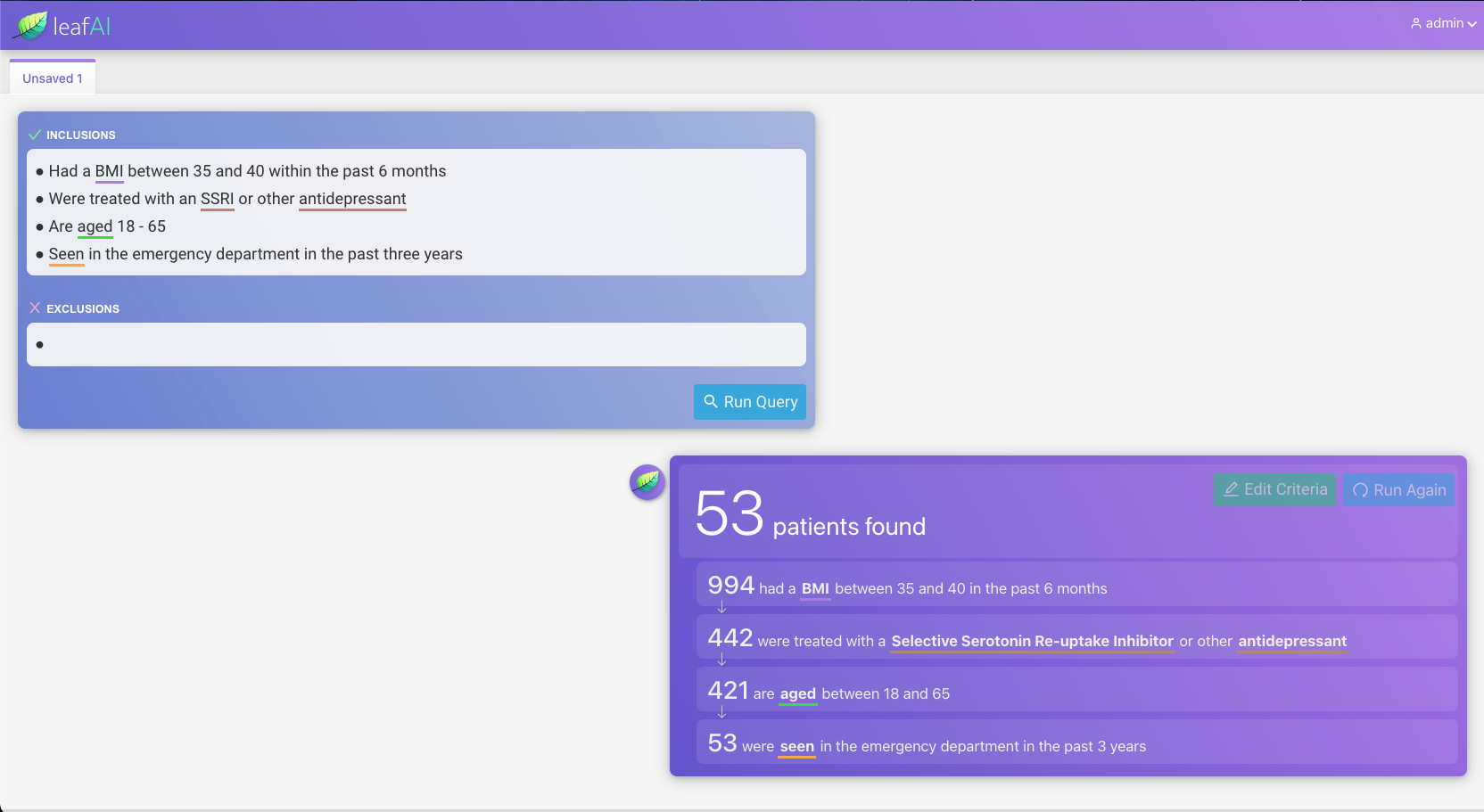}  
\caption{Example screenshot of the LeafAI web application, which is currently in development.}
\label{fig_leafai_screenshot}
\end{figure}

We are actively developing a web-based user interface for LeafAI, shown in Figure \ref{fig_leafai_screenshot}. In future work, we will deploy a prototype of the tool and evaluate user feedback and system performance. The LeafAI web application will provide rapid feedback to users explaining its search strategies, and allow users to override system-reasoned concepts and edit or add their own. Additionally, we intend to explore the adaptation of our logical form-based query generation methods to general-purpose question answering and querying systems such as FHIR endpoints.

\section*{Conclusion}
\label{sec:conclusion}

This study introduced LeafAI, an NLP-based system leveraging deep learning and KB which can automatically generate queries for cohort discovery on virtually any clinical data model. Using an OMOP database representing the entire patient population of our institution, we demonstrated that LeafAI rivals the  performance of a human programmer in identifying eligible research candidates. As future work we will deploy LeafAI for our research community, obtaining their feedback and iteratively improving the  tool.

%TC:ignore 
\section*{Funding} 

This study was supported in part by the National Library of Medicine under Award Number R15LM013209 and by the National Center for Advancing Translational Sciences of National Institutes of Health under Award Number UL1TR002319.

\section*{Author contributions statement}

ND is the developer of LeafAI and wrote the majority of the manuscript. BH and WZ annotated the LLF dataset and contributed to the annotation schema. KL served as the human database programmer, and NK and RH advised on study design. OU and MY advised on strategies for query generation and NLP architectures. All authors contributed to the interpretation of the data, manuscript revisions, and intellectual value to the manuscript.

\section*{Competing interests}

The authors declare no competing interests.

\section*{Data Availability}
The annotation data for the LCT corpus is available at \url{https://github.com/uw-bionlp/clinical-trials-gov-data/tree/master/ner/nic}. The annotation data for the LLF corpus is available at \url{https://github.com/ndobb/clinical-trials-seq2seq-annotation}.

\bibliography{main}

\section*{Appendix}
\label{sec:appendix}

\subsection*{Leaf Logical Forms corpus}

We developed annotation guidelines for the LLF corpus using a simplification of entities and relations from the preceding LCT corpus \cite{dobbins2022leaf}. Generally speaking, LCT \textit{entities} correspond to logical form \textit{functions}, while LCT \textit{relations} correspond to logical form \textit{predicates}. For example, the LCT \textit{Condition} entity has a corresponding \textit{cond()} function, while the \textit{Num-Filter} relation has a corresponding \textit{.num\_filter()} predicate. The LLF corpus contains 44 total functions and 20 total predicates. The LLF annotation guidelines can be found at \url{https://github.com/ndobb/clinical-trials-seq2seq-annotation/wiki}.

We also hypothesized that the performance of predicting logical forms could likely be improved by replacing "raw" tokens in each eligibility criteria with corresponding logical form names derived from named entities from the LCT corpus. For example, given the eligibility criterion: \\

"\textit{Diabetics who smoke}", \\

\noindent we would replace the named entities for "Diabetics" and "smoke": \\ 

\textit{cond("Diabetics") who obs("smoke")} \\

\noindent using \textit{Condition} and \textit{Observation} annotations in the LCT corpus. We call this substituted text an "augmented" eligibility criteria. The augmented criteria syntax reshapes named entities to more closely resemble expected logical form syntax and allows us to leverage the LCT corpus for logical form transformation.

\begin{figure}[H]
  \centering
  \includegraphics[scale=0.7]{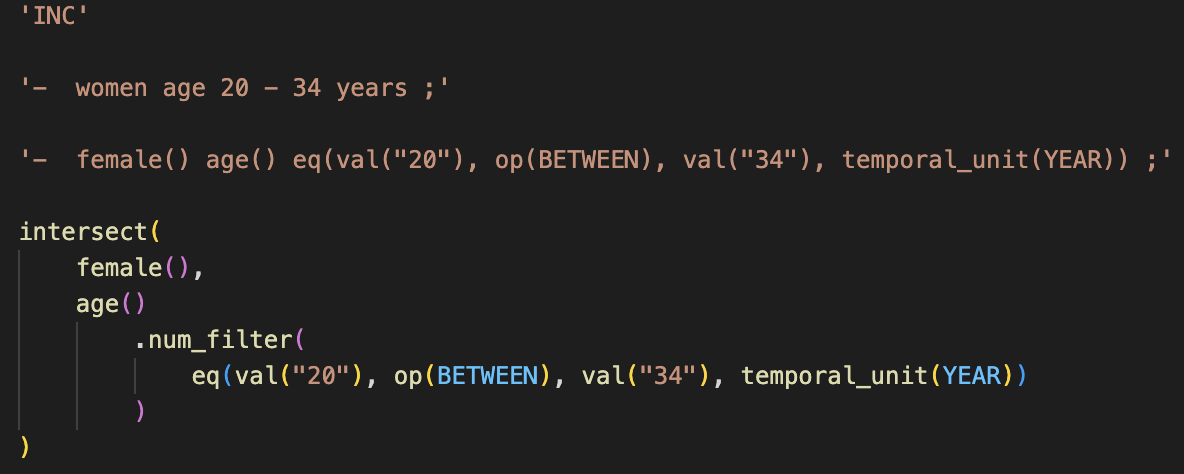}  
  \caption{An example LLF corpus annotation. The annotation file is saved in JavaScript (.js) format, which enables syntax highlighting and validation to assist annotators. Whether a given criterion was an inclusion or exclusion criteria is indicated at the top, followed by the original raw text, then augmented text. The final annotated logical forms are shown last.}
\label{fig_annotation_example}
\end{figure}

Creation and annotation of the LLF corpus proceeded in the following steps:

\begin{enumerate}
    \item We randomly chose 2,000 lines of eligibility criteria from the LCT corpus, limited to only criteria which included at least one named entity and which were not annotated as hypothetical criteria. 30\% of the 2,000 lines (600) were randomly chosen among lines with particularly complex entity and relation types, such as \textit{If-Then}, \textit{Before-After}, \textit{Contraindication}, etc.
    \item  Each annotation file consists of the text "EXC" if exclusion or "INC" if inclusion (line 1), an original "raw" eligibility criteria (line 3), an augmented eligibility criteria (line 5), and an (initially blank) expected logical form equivalent to annotate (line 7). An example annotation is shown in Figure \ref{fig_annotation_example}.
    \item 3 informatics PhD students knowledgeable about the task met weekly for 2 months to review annotations and resolve difficult cases together. Annotators were initially trained on 20 lines of eligibility criteria which included 31 different function types.
    \item After training, each annotator was assigned a batch of 100 sentences (one per file) and tasked with writing a logical form version of each.
    \item After each batch was completed, we executed a quality control script to parse each logical form annotation to ensure consistency. Any syntax errors were reported to and corrected by the annotators.
    \item Annotators received additional batches of files to annotate until all 2,000 single-annotated annotations had been completed.
    \item As only 20 (1\%) of the 2,000 files in the corpus were triple-annotated and to measure consistency, after annotation we randomly selected 10 files from each annotator and assigned 5 each for the other annotators to review, for 30 in total. Annotators ranked each annotation in a binary fashion, with 1 indicating a completely correct annotation, and 0 indicating the presence of a possible mistake. Of the 30 randomly chosen annotations, 28, or 93\%, were agreed upon by the annotators as correct.
\end{enumerate}

The pair-wise mean inter-annotator agreement by BLEU score was 82.4\%. After annotations were completed, we experimented with predicting logical forms by fine-tuning T5 \cite{raffel2020exploring} Seq2Seq models. The T5 architecture and pre-trained models are widely used for and achieve at or near state-of-the-art for many machine translation and semantic parsing tasks. 

Following earlier work on task-oriented dialog semantic parsing structures in the domain of digital assistants, we also experimented using various alternative input-output syntax styles from our original logical forms:

\begin{enumerate}
    \item \textbf{Shift-Reduce}. Einolghozatic \textit{et al.} \cite{einolghozati2019improving} used square brackets instead of parentheses and blank spaces instead of commas. We followed Rongali \textit{et al's} suggestion to add a trailing repeat of function names to improve performance.
    \item \textbf{Pointer}. Rongali \textit{et al.} \cite{rongali2020don} found that replacing input tokens with "$@ptr_{index}$", where \textit{index} corresponds to a token's sequential position in the input text improved performance in their semantic parsing task. We modified this approach by omitting the characters "ptr" and using the sequential position of the quoted span as our index rather than individual token positions.
\end{enumerate}

We used a randomly sorted 70/20/10 train/test/validation split of the LFF corpus to fine-tune the pretrained T5$_{base}$ model using combinations of these syntax styles. We call our gold standard annotated logical form syntax "Standard" style. Example inputs, outputs, and training results are shown in Table \ref{tbl_llf_corpus}. 

\begin{table}[t!]
    \footnotesize
    \centering
    \def\arraystretch{1.4}
\definecolor{gray}{RGB}{230,230,230}
\begin{tabular}{m{2.5cm} l l l l}
    \toprule
    \textbf{Syntax Style} & \textbf{Example Input} & \textbf{Example Logical Form} & \textbf{BLEU} & \textbf{ROUGE-L} \\
    \midrule
    \arrayrulecolor{gray}
    Raw-text→ Standard       
       & Diabetics who smoke                     
       & $\makecell[cl]{intersect(\\\mathrm{\ \ \ }cond("Diabetics"), \\\mathrm{\ \ \ }obs("smoke")\\)}$
       & 78.7
       & 79.1 \\
    \midrule
    Standard  
       & cond("Diabetics") who obs("smoke")           
       & $\makecell[cl]{intersect(\\\mathrm{\ \ \ }cond("Diabetics"), \\\mathrm{\ \ \ }obs("smoke")\\)}$
       & \textbf{93.5}
       & \textbf{92.3} \\
    \midrule
    Standard+ Pointer
       & cond(@1) who obs(@2)                          
       & $\makecell[cl]{intersect(\\\mathrm{\ \ \ }cond(@1), \\\mathrm{\ \ \ }obs(@2)\\)}$
       & 93.3
       & 91.2 \\
    \midrule
    Shift-Reduce              
       & [cond "Diabetics" cond] who [obs "smoke" obs] 
       & $\makecell[cl]{[intersect\\\mathrm{\ \ \ }[cond\mathrm{\ }"Diabetics"\mathrm{\ }cond]\\\mathrm{\ \ \ }[obs\mathrm{\ }"smoke"\mathrm{\ }obs]\\ intersect]}$
       & 89.8
       & 91.7 \\
    \midrule
    Shift-Reduce+ Pointer     
       & [cond @1 cond] who [obs @2 obs]               
       & $\makecell[cl]{[intersect\\\mathrm{\ \ \ }[cond\mathrm{\ }@1\mathrm{\ }cond]\\\mathrm{\ \ \ }[obs\mathrm{\ }@2\mathrm{\ }obs]\\ intersect]}$
       & 89.4
       & 90.4 \\
    \bottomrule       
\end{tabular}
    \caption{Example inputs and logical form syntax styles with fine-tuning performance results using the T5$_{base}$ model.}
    \label{tbl_llf_corpus}
\end{table} 

We found that our Standard logical forms achieved the highest performance using both BLEU \cite{lin2004rouge} and ROUGE-L \cite{ callison2006re} scores, two commonly used metrics in measuring Seq2Seq performance. Replacing raw tokens with function names corresponding to named entities also significantly improved performance (+14.7\%, comparing raw text to Standard input styles), demonstrating that leveraging the LCT corpus to generate augmented text achieved relatively high performance (> 93\% BLEU score) for this task. As it was the highest-performing syntax style and also the most straightforward to parse, we chose to use the Standard logical form style as our IR for LeafAI.

\subsection*{Evaluation}

While we found our Standard logical form syntax to achieve in our evaluation with T5, we acknowledge that 20 of the 2,000 total criteria in the LLF corpus were used for annotator training, which may not be representative of the corpus as a whole. However, of the 30 randomly chosen annotations, the annotators found 28 to be correct, or 93\%. This value was very close to our best performing T5 evaluation (93.5\% BLEU score) and somewhat higher than the inter-annotator agreement on the training annotations (82.4\% BLEU score). Additionally, all annotators were informatics PhD students knowledgeable about the task and goal, and met weekly to resolve difficult cases together. So while only 1\% of the corpus was triple-annotated, we believe the annotation was performed reasonably consistently.

\subsection*{Limitations}

While we found our Standard logical form syntax to achieve in our evaluation with T5, we acknowledge that 20 of the 2,000 total criteria in the LLF corpus were used for annotator training, which may not be representative of the corpus as a whole. However, of the 30 randomly chosen annotations, the annotators found 28 to be correct, or 93\%. This value was very close to our best performing T5 evaluation (93.5\% BLEU score) and somewhat higher than the inter-annotator agreement on the training annotations (82.4\% BLEU score). Additionally, all annotators were informatics PhD students knowledgeable about the task and goal, and met weekly to resolve difficult cases together. So while only 1\% of the corpus was triple-annotated, we believe the annotation was performed reasonably consistently.

%TC:endignore 

\end{document}